\documentclass[lettersize,journal]{IEEEtran}
\usepackage{amsmath,amsfonts}
 \usepackage{amssymb}
\usepackage{algorithm,algorithmicx,algpseudocode}
\usepackage{multirow}
\usepackage{array}
\usepackage[caption=false,font=normalsize,labelfont=sf,textfont=sf]{subfig}
\usepackage{textcomp}
\usepackage{stfloats}
\usepackage{url}
\usepackage{verbatim}
\usepackage{graphicx}
\usepackage{cite}
\usepackage{booktabs} 
\usepackage{hyperref}
\usepackage{xcolor}
\usepackage{nicefrac}       
\usepackage{bm}     
\begin{document}

\renewcommand{\algorithmicrequire}{\textbf{Input:}}
\renewcommand{\algorithmicensure}{\textbf{Output:}}

\title{Gradient-Free Adversarial Purification with Diffusion Models}

\author{Xuelong Dai, Dong Wang, Xiuzhen Cheng~\IEEEmembership{Fellow,~IEEE}, Bin Xiao~\IEEEmembership{Fellow,~IEEE} 
\thanks{Bin Xiao is the corresponding author.}
\thanks{Xuelong Dai and Xiuzhen Cheng are with the School of Artificial Intelligence, Shandong University, Jinan, China. (email: daixuelong@sdu.edu.cn, xzcheng@sdu.edu.cn)}
\thanks{Dong Wang and Bin Xiao are with the Department of Computing, The Hong Kong Polytechnic
University, Hong Kong. (email: dong-comp.wang@connect.polyu.hk, b.xiao@polyu.edu.hk)}
}

\markboth{IEEE Transactions on Information Forensics and Security}%
{Dai \MakeLowercase{\textit{et al.}}: GFAP}


\maketitle

\begin{abstract}

Adversarial training and adversarial purification are two widely used defense strategies for enhancing model robustness against adversarial attacks. However, adversarial training requires costly retraining, while adversarial purification often suffers from low efficiency. More critically, existing defenses are primarily designed under the perturbation-based adversarial threat model, which is ineffective against recently introduced unrestricted adversarial attacks.
In this paper, we propose an effective and efficient defense framework that counters both perturbation-based and unrestricted adversarial attacks. Our approach is motivated by the observation that adversarial examples typically lie near the decision boundary and are highly sensitive to pixel-level perturbations. To address this, we introduce adversarial anti-aliasing, a preprocessing technique that mitigates adversarial noise by reducing the magnitude of pixel-level perturbations. In addition, we propose adversarial super-resolution, which leverages prior knowledge from clean datasets to benignly restore high-quality images from adversarially degraded ones. Unlike image synthesis methods that generate entirely new images, adversarial super-resolution focuses on image restoration, making it more suitable for purification. Importantly, both techniques require no additional training and are computationally efficient since they do not rely on gradient computations.
To further improve robustness across diverse datasets, we introduce a contrastive learning–based adversarial deblurring fine-tuning method. By incorporating adversarial priors during fine-tuning on the target dataset, this method enhances purification effectiveness without the need to retrain diffusion models.
Extensive experiments demonstrate that our defense consistently outperforms state-of-the-art adversarial purification methods against both perturbation-based and unrestricted adversarial attacks, achieving superior standard accuracy, robust accuracy, and time efficiency. Moreover, our defense framework is highly flexible and can be seamlessly integrated with existing defense strategies for even stronger purification.

\end{abstract}


\section{Introduction}

\IEEEPARstart{D}{eep} learning models have demonstrated remarkable performance across various tasks \cite{he2016deep,liu2021swin,Xiang_2021_ICCV}. With the rapid advancement and widespread deployment of these models, their security and robustness are garnering increasing attention from the research community.

It is widely recognized that deep learning models are highly vulnerable to adversarial attacks \cite{madry2017towards,carlini2017towards}. These attacks are performed by adding imperceptible perturbations to clean images. The perturbed images, known as adversarial examples, can deceive trained deep learning classifiers with high confidence while appearing natural and realistic to human observers. To mitigate adversarial attacks and ensure the stability of deep learning models, adversarial training \cite{madry2017towards,gowal2021improving} has been developed. This approach defends against adversarial attacks by training the classifier with adversarial examples. However, adversarial training suffers from two major limitations. First, it incurs substantial computational overhead. Unlike standard training, which requires only a single backward propagation per step, adversarial training necessitates generating adversarial examples through multiple rounds of gradient backpropagation. As a result, the overall training cost is several times higher than that of standard training. Second, adversarial training exhibits poor transferability to unseen attacks. It is typically effective only against the specific attacks included during training. To mitigate newly emerging adversarial threats, the model must undergo additional retraining, further compounding its inefficiency.

Adversarial purification \cite{samangouei2018defense,wongfast} seeks to remove adversarial perturbations from adversarial examples and restore clean images by introducing a pre-processing model before the target model. Unlike adversarial training, purification does not require retraining the target model and thus significantly reduces training overhead. Early approaches employed denoising networks or generative models, such as generative adversarial networks (GANs). However, these methods often suffered from overfitting or the limited generative capacity of the models.
With the recent advances in diffusion models \cite{dhariwal2021diffusion,rombach2022high}, diffusion-based adversarial purification \cite{nie2022DiffPure,song2024mimicdiffusion,chen2024diffilter} has demonstrated strong defense performance by effectively recovering adversarial examples to clean images. These methods leverage the reverse generation process of diffusion models, which gradually removes both Gaussian noise introduced in the forward process and adversarial perturbations. Diffusion-based purification is particularly suitable for this task, as the denoising mechanism inherent in diffusion training and sampling can neutralize adversarial noise through injected Gaussian perturbations.
Despite their effectiveness, diffusion-based purification methods face a major limitation: the sampling process is computationally expensive and time-consuming, making them impractical for real-time defense scenarios.

Diffusion models also facilitate stronger unrestricted adversarial attacks \cite{chen2023advdiffuser,dai2023advdiff,chen2023content}. These unrestricted adversarial examples (UAEs) are generated through a reverse-generation process guided by adversarial objectives. Unlike traditional perturbation-based attacks, UAEs demonstrate superior effectiveness against current defenses due to their fundamentally different threat model. They present a significant challenge to the development of deep learning models, as they are directly generated by diffusion models rather than being crafted through small perturbations on benign data. As out-of-distribution samples, UAEs can effectively bypass neural networks trained on standard data distributions. Alarmingly, existing defenses provide only limited coverage against these attacks, highlighting the urgent need for new defense strategies specifically targeting UAEs.

\begin{figure*}[t]
   \centering
     \includegraphics[width=0.8\linewidth]{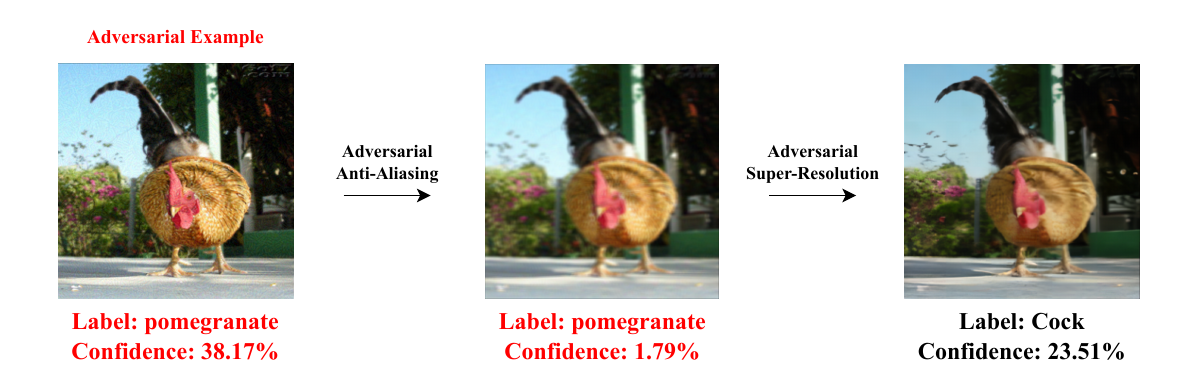}
      \caption{\textbf{The proposed adversarial defense pipeline.} We give an adversarial example of ``cock'' class with AutoAttack $\ell_\infty=8/255$ on ImageNet dataset. Adversarial anti-aliasing aims to eliminate adversarial perturbations, while adversarial super-resolution seeks to restore benign images from blurred adversarial examples using prior knowledge from the clean dataset.}
   \label{fig:1}
    \vspace{-2mm}
   \end{figure*}

In this paper, we propose an effective adversarial defense method that detects both perturbation-based adversarial examples and unrestricted adversarial examples. To achieve the defense objective, we locate and utilize the common characteristic of these two types of attacks that both adversarial examples are generated close to the decision boundary for minimal perturbations, which makes these adversarial examples susceptible to changes in pixels. 

Our defense adopts a zero-shot adversarial purification strategy by extracting the semantic shape of images while discarding fine-grained details with adversarial perturbations, as illustrated in Figure~\ref{fig:1}. Specifically, we employ adversarial anti-aliasing with specialized filters to suppress adversarial perturbations, followed by adversarial super-resolution, which reconstructs high-quality details from pre-trained clean super-resolution diffusion models. Both methods are time-efficient and require no modifications to the original models.
To further enhance performance across diverse target datasets, we introduce a contrastive learning-based adversarial deblurring fine-tuning method. This approach leverages our adversarial anti-aliasing to construct paired low-quality and high-quality images, enabling the model to learn dataset-specific priors and thereby improve purification effectiveness on various trained target models. Additionally, we incorporate adversarial contrastive learning into the deblurring task, equipping the diffusion model with adversarial-aware priors to strengthen defense robustness.
We validate the effectiveness of our defense by applying the purified adversarial examples as inputs to existing adversarial purification methods, demonstrating substantial improvements in their defense performance. Experiments across multiple datasets confirm that our approach not only outperforms state-of-the-art adversarial defenses in purification quality but also achieves an average $\times 44$ speed-up in time efficiency compared to existing diffusion-based defenses.

Our contributions are summarized as follows:
\begin{itemize}
    \item We propose a novel adversarial defense capable of countering both perturbation-based adversarial examples and unrestricted adversarial examples, addressing the current gap in effective defenses against unrestricted adversarial attacks.
    \item We introduce various zero-shot and gradient-free defense strategies that preserve the semantic information of adversarial examples while eliminating adversarial modifications with theoretical supports. These strategies include adversarial anti-aliasing for ``semantic'' extraction and adversarial super-resolution for incorporating benign priors and recovering benign details from adversarial examples.
    \item We present a contrastive learning-based adversarial deblurring fine-tuning method, which learns dataset-specific and adversarial-aware priors to further enhance the performance of our adversarial purification across diverse target datasets.
    \item We conduct extensive experiments on various datasets against adaptive adversarial attacks. The results demonstrate the effectiveness of our proposed defense method compared to state-of-the-art adversarial defenses. Moreover, anti-aliased and upscaled adversarial examples effectively integrate with existing diffusion-based adversarial purification, validating the usability and scalability of our approach.
\end{itemize}

\section{Background}

\subsection{Adversarial Training}

Adversarial training (AT) is among the most practical and widely studied methods for enhancing model robustness against adversarial attacks \cite{goodfellow2014explaining}. The central idea is to expose the model to both benign and adversarial examples during training, enabling it to learn representations that are resilient to malicious perturbations. Madry et al. \cite{madry2017towards} provided a comprehensive formulation of adversarial training as a robust optimization problem, establishing a strong theoretical foundation. Building on this, Kannan et al. \cite{kannan2018adversarial} improved the performance of AT by encouraging similarity in the logits of paired examples, thereby stabilizing learning.
Despite these advances, a key limitation of adversarial training lies in its trade-off between robustness and standard accuracy: while it enhances robustness, the introduction of adversarial examples during training often reduces clean accuracy. Zhang et al. \cite{zhang2019theoretically} addressed this issue by designing specialized loss functions that explicitly balance clean and robust accuracy, leading to significant improvements in both. Nevertheless, robustness against unseen or out-of-distribution attacks remains a major challenge for traditional AT. To overcome this limitation, Song et al. \cite{songimproving} incorporated domain adaptation techniques to enhance the generalization ability of adversarially trained models, while Gowal et al. \cite{gowal2021improving} showed that leveraging generated or even unrealistic synthetic data can further strengthen robustness.
Another well-documented issue in adversarial training is robust overfitting, where models overfit to the specific adversarial examples seen during training, thus weakening generalization to novel attacks. Rebuffi et al. \cite{rebuffi2021data} addressed this by incorporating augmented data, which increased diversity and mitigated overfitting. In addition to enriching training data, refining the training objective has proven effective in improving generalization. Wu et al. \cite{wu2020adversarial} proposed Adversarial Weight Perturbation (AWP), which extends perturbations not only to the inputs but also to the network weights, thereby improving robustness against a broader range of attacks. More recently, Jin et al. \cite{jin2023randomized} introduced stochasticity into adversarial training by applying a robustness loss function over randomized weights, achieving gains in both clean and robust accuracy.

\subsection{Adversarial Purification}

Adversarial purification aims to remove adversarial perturbations from adversarial examples without requiring the retraining of deep learning models. Early approaches \cite{guo2017countering,liu2019feature} adopted image transformation techniques like JPEG compression to suppress adversarial noise, but these methods were effective only against small perturbations and lacked robustness against stronger attacks. To improve purification strength, subsequent works leveraged the generative capabilities of deep generative models. Samangouei et al. \cite{samangouei2018defense} employed Generative Adversarial Networks (GANs) to sample clean images close to adversarial counterparts by training a defense GAN. Score-based generative models \cite{songscore} also demonstrated strong purification capabilities. For example, Yoon et al. \cite{yoon2021adversarial} introduced an Energy-Based Model combined with Markov Chain Monte Carlo (MCMC) sampling for purification, further improving sampling efficiency through denoising score matching.
With the emergence of diffusion models \cite{ho2020denoising,dhariwal2021diffusion}, a new paradigm of adversarial purification was introduced. Nie et al. \cite{nie2022DiffPure} showed that diffusion-based purification significantly outperforms prior approaches by exploiting the reverse generation process. Their method adopting the diffusion reverse sampling process with optimal diffusion steps to remove the adversarial perturbations within the sampled Gaussian noise. However, finding the optimal generation steps for diffusion-based adversarial purification remains challenging. To address this limitation, Lee et al. \cite{lee2023robust} proposed a gradual noise-scheduling strategy with tailored forward-step schedules, improving purification stability. Bai et al. \cite{baidiffusion} introduced contrastive guidance to refine the sampling process and enhance purification performance. Lin et al. \cite{linadversarial} incorporated adversarial training into diffusion models, further strengthening their defensive capability.
Another challenge lies in the vulnerability of diffusion’s reverse generation process to adversarial images. Wang et al. \cite{wang2022guided} tackled this problem using a guided diffusion approach, which eliminates the need for optimal step selection by initiating purification from random Gaussian noise rather than adversarial inputs. Building on this, Song et al. \cite{song2024mimicdiffusion} introduced proposed two guidance, long-range and short-range guidance, to enhance the adversarial purification with the guided diffusion model.

Despite these advances, existing purification methods face two major limitations: (1) low time efficiency, particularly for gradient-based guidance methods, and (2) poor effectiveness against unrestricted adversarial examples (UAEs). These challenges highlight the need for more efficient and generalizable adversarial purification techniques.

\begin{figure}[t]
   \centering
     \includegraphics[width=1.0\linewidth]{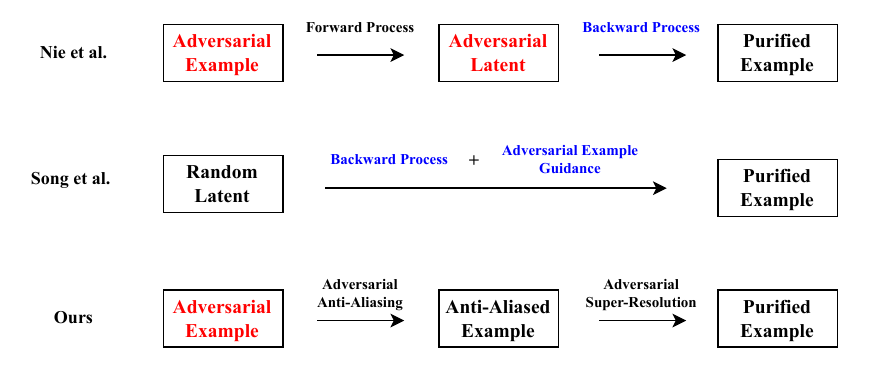}
      \caption{\textbf{The comparisons of state-of-the-art diffusion-based adversarial purification pipelines.} We mark the defense process in \textcolor{blue}{blue} to represent time-consuming approaches. We use \textcolor{red}{red} font to indicate non-purified adversarial examples with adversarial perturbations.}
   \label{fig:pip}
    \vspace{-5mm}
   \end{figure}
\section{Preliminary}

\subsection{Threat Model}

Adversarial examples conduct attacks by fooling the target model's classification result. Considering the untargeted attack scenario, the perturbation-based adversarial examples are defined as:

\begin{equation}
   A_{\text{AE}} \triangleq \{\bm{x}_\text{adv} = \bm{x} + \bm{\delta} |y \neq f(\bm{x}), \bm{x} \in \mathcal{D}, |\bm{\delta}| \leq \kappa\}
\end{equation}

where $\bm{\delta}$ is the adversarial perturbation, $f(\cdot)$ is the target model, $\mathcal{D}$ is the clean dataset, and $\kappa$ is the perturbation norm constraint.

These adversarial examples are generated by adding the perturbations to the clean images. However, such perturbations can degenerate the image quality. By utilizing the generation models, Song et al. \cite{song2018constructing} presented unrestricted adversarial examples by directly generating adversarial examples with the generation tasks, which can be formulated as:

\begin{equation}
   A_{\text{UAE}} \triangleq \{\bm{x}_\text{adv} \in \mathcal{G}(\bm{z}_{\text{adv}},y)|y \neq f(\bm{x})\}
\end{equation}
where $\mathcal{G}$ is the generation model, $\bm{z}_{\text{adv}}$ is the latent code for generation.

These two adversarial examples are generated with different threat models. However, they both can successfully conduct attacks against the given target model. A robust defense method should be able to defend against these attacks simultaneously.

\subsection{Diffusion-Based Adversarial Purification}

The diffusion model \cite{ho2020denoising} learns to recover the image from the denoising-like process, i.e., \textit{reverse generation process}. The reverse generation process takes $T$ time steps to obtain a sequence of noisy data $\{\bm{x}_{T-1}, \dots, \bm{x}_{1}\}$ and get the data $\bm{x}_0$ at the last step. Specifically, it can be formulated as:
\begin{equation}
    p_\theta(\bm{x}_{t-1}|\bm{x}_t) = \mathcal{N}(\bm{x}_{t-1}:\mu_{\theta}(\bm{x}_t,t),\Sigma_{\theta}(\bm{x}_t,t)) 
\end{equation}

The \textit{forward diffusion process} is where we iteratively add Gaussian noise to the data for training the diffusion model to learn $p_\theta(\bm{x}_{t-1}|\bm{x}_t)$. It is defined as:
\begin{equation}
    q(\bm{x}_t|\bm{x}_{t-1})=\mathcal{N}(\bm{x}_t:\sqrt[]{\sigma_t}\bm{x}_{t-1},(1-\sigma_t)\textbf{I} ) 
\end{equation}
where $\sigma$ is the noise schedule.

Nie et al. \cite{nie2022DiffPure} attempted to find the optimal $t^*$ where it satisfy that:
\begin{align}
    \bm{x}_{t^*} &= \sqrt[]{\sigma_{t^*}}\bm{x}_\text{adv} + \sqrt[]{1-\sigma_{t^*}} \bm{\varepsilon} \\ \notag
            &= \sqrt[]{\sigma_{t^*}}(\bm{x}+\bm{\delta}) + \sqrt[]{1-\sigma_{t^*}} \bm{\varepsilon}
\end{align}
where $\bm{\varepsilon}$ is the Gaussian noise $\bm{\varepsilon}\sim \mathcal{N}(0,\textbf{I})$. After we obtain the optimal $t^*$, we can utilize the reverse generation process over $\bm{x}_\text{adv}$ to recover the clean $\bm{x}$.

\begin{figure*}[t]
   \centering
     \includegraphics[width=0.8\linewidth]{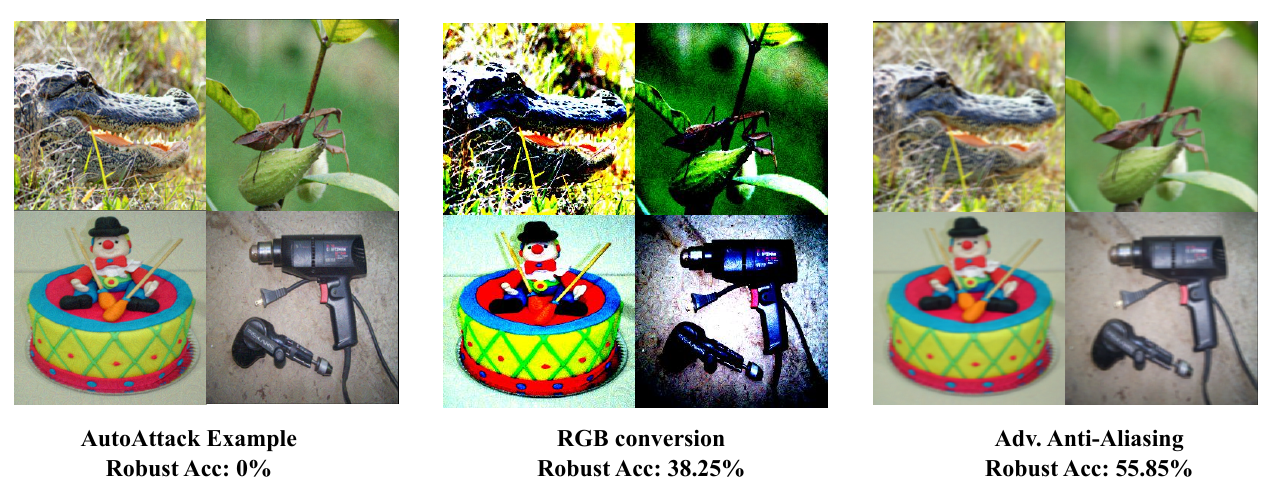}
      \caption{\textbf{The vulnerability of adversarial examples to the changes in pixels.} The RGB conversion is performed by converting the images to RGB space after the ImageNet normalization and achieves 38\% robust accuracy. The proposed adversarial anti-aliasing is more effective while preserving the image quality.}
     \vspace{-2mm}  
   \label{fig:2}
   \end{figure*}

Song et al. \cite{song2024mimicdiffusion} utilized the whole reverse generation process with $T$ time step; they used adversarial sample $\bm{x}_\text{adv}$ as guidance rather than an intermediate time step state. At each time step $t$, the guidance is added to the $\bm{x}_t$ after the original reverse generation process and can be formulated as:
\begin{equation}
     \nabla_{\bm{x}} \log p(\bm{x}_\text{adv}|\bm{x}_t;t) = -R_t\nabla_{\bm{x}_t}d(\bm{\hat{x}}_t,\bm{x}_\text{adv})
\end{equation}
where $R_t$ is the scale factor at $t$ time step, $d(\cdot)$ is the distance measurement, and $\bm{\hat{x}}_t$ is the estimation for $\bm{x}_0$ at $t$ time step. The $\bm{\hat{x}}_t$ is defined as:
\begin{equation}
    \bm{\hat{x}}_t =\frac{\bm{x}_t - \sqrt[]{1-\sigma_{t}}s_\theta(\bm{x}_t)}{\sqrt[]{\sigma_t}}
\end{equation}
where the  $s_\theta$ known score function is defined as \cite{songscore}.

\section{Gradient-Free Adversarial Super-Resolution Purification} 
\label{Puri}
\subsection{Motivation}


With the advancement of diffusion models, diffusion-based adversarial purification has emerged as a leading approach for adversarial defenses. However, current methods still face significant challenges that impact their effectiveness. Figure \ref{fig:pip} illustrates typical diffusion-based purification pipelines from state-of-the-art methods. Nie et al. \cite{nie2022DiffPure} achieved purification by utilizing the adversarial latent generated by the forward process of adversarial examples. Unfortunately, this approach can introduce adversarial perturbations into the purified examples, as these perturbations persist in the adversarial latent. Song et al. \cite{song2024mimicdiffusion} sought to mitigate the impact of adversarial perturbations by using random latents, employing adversarial examples solely as guidance. However, this method requires gradient calculations at each step of the reverse process, making it computationally intensive. Consequently, achieving both \textbf{time-efficient} and \textbf{perturbation-isolated} diffusion-based adversarial purification remains a challenge. Furthermore, existing defenses fail to defend against the recently proposed unrestricted adversarial attacks.

\begin{figure*}[t]
   \centering
     \includegraphics[width=1.0\linewidth]{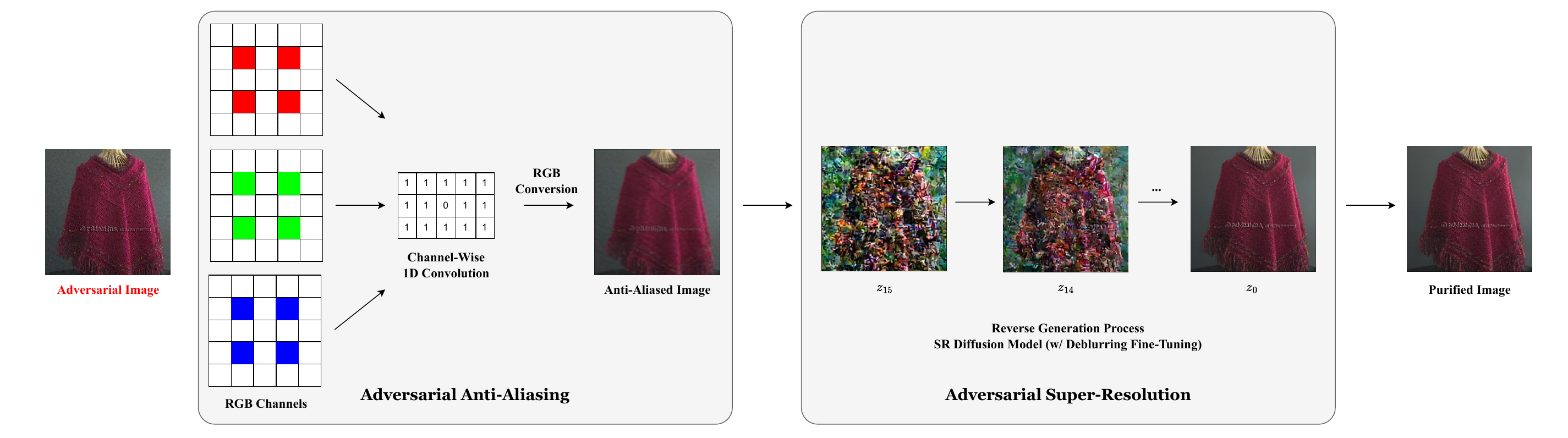}
      \caption{\textbf{The proposed adversarial purification pipeline.} Our proposed adversarial purification framework consists of two defense stages. In the first stage, we apply adversarial anti-aliasing as a pre-processing technique to reduce the magnitude of adversarial perturbations while preserving the semantic structure of the image through controlled pixel degradation. Specifically, we perform channel-wise 1D convolution on each RGB channel of the adversarial input, using in-place operations to preserve the original image resolution. The processed RGB channels are then combined to form the anti-aliased image. In the second stage, we apply adversarial super-resolution using a pretrained diffusion model. The purification is performed via the reverse generation process of the super-resolution (SR) diffusion model, which enhances image quality while mitigating residual adversarial noise. To further improve robustness, we introduce deblurring fine-tuning of the SR model, as detailed in Section \ref{Debl}. The final purified image is obtained as the output of this SR diffusion process.}
   \label{fig:aa} 
    \vspace{-2mm}
   \end{figure*}
   
\subsection{Perturbation-Isolated Adversarial Purification}
Perturbation-based adversarial examples are precisely calculated based on the gradient of the loss function, whereas unrestricted adversarial examples are sampled near the decision boundary. Despite employing different threat models, both types of attacks produce adversarial examples that are sensitive to pixel changes. Since adversarial examples are designed to be imperceptible compared to clean images, the semantic shapes of objects within the images should correspond to their original labels. Therefore, our defense strategy focuses on extracting the semantic shapes from the adversarial examples and eliminating the adversarial pixel-level details, as illustrated in Figure \ref{fig:aa}. 

\subsubsection{Adversarial Anti-Aliasing}

To achieve effective defenses against both unrestricted and perturbation-based adversarial attacks, it is essential to address their common characteristics. One critical factor is the value range of images: a valid RGB value is an integer between 0 and 255. However, the modifications introduced by various adversarial attacks are often performed using non-integer data types for gradient calculations. These modifications can become ineffective when transformed back to the RGB image format. Figure \ref{fig:2} supports our findings, showing that approximately 38\% of adversarial examples from AutoAttack fail with the combinations of RGB conversions and image normalization for deep-learning models. The reasons for this phenomenon could be that adversarial examples are typically located near the decision boundary and are sensitive to pixel changes. However, simple RGB conversion can be effectively compromised by adaptive attacks \cite{athalye2018obfuscated}. Therefore, in this paper, we aim to propose more effective transformations.

Anti-aliasing is a straightforward, zero-shot method for smoothing image details, including adversarial perturbations \cite{liang2018detecting,vasconcelos2021impact}. Unlike previous works, we have found that anti-aliasing with non-square filters is particularly effective against adversarial attacks while preserving clean accuracy. Additionally, using the average value from neighboring pixels, excluding the original pixel, has also proven effective. This is because adversarial perturbations are calculated on a pixel-wise basis and are sensitive to pixel changes. These two approaches greatly enhance the effectiveness of anti-aliasing. Even with simple anti-aliasing, we achieve moderate defense performance, underscoring the effectiveness of our approach. We give the details on the implementation of proposed adversarial anti-aliasing in Figure \ref{fig:aa}. We use 1D convolution on each RBG channel of the adversarial image and clip the additional dimensions according to the input resolution. After that, we convert the anti-aliased image tensor to the RGB image for adversarial super-resolution.

To maintain the resolution of the output image, we use padding, which is calculated as follows:
\begin{equation}
    R_{out}=\left \lfloor R_{in}+2 \times \text{Padding}-\text{filter\_size} \right \rfloor 
\end{equation}
where $R$ is the shape of the data. We use stride $=1$.

The proposed adversarial anti-aliasing utilizes the mean value from a pixel's neighborhood and can be seen as a direct method for reducing adversarial perturbations. It is well-known in image classification that the areas contributing most to classification results are typically located at the center of the main object, where they are rich in textures. Consequently, the anti-aliasing filter can effectively neutralize adversarial perturbations in these regions, which often exhibit significant changes in RGB values and gradients of the classification loss function \cite{jiang2013salient}, especially in high-resolution images. By applying blurring, the proposed adversarial anti-aliasing method can significantly decrease the magnitude of adversarial perturbations. The worst case of the proposed method occurs in simple backgrounds, such as the sky. However, these areas generally have minimal impact on the final classification results.

\subsubsection{Adversarial Super-Resolution}

\begin{figure*}[t]
   \centering
     \includegraphics[width=0.7\linewidth]{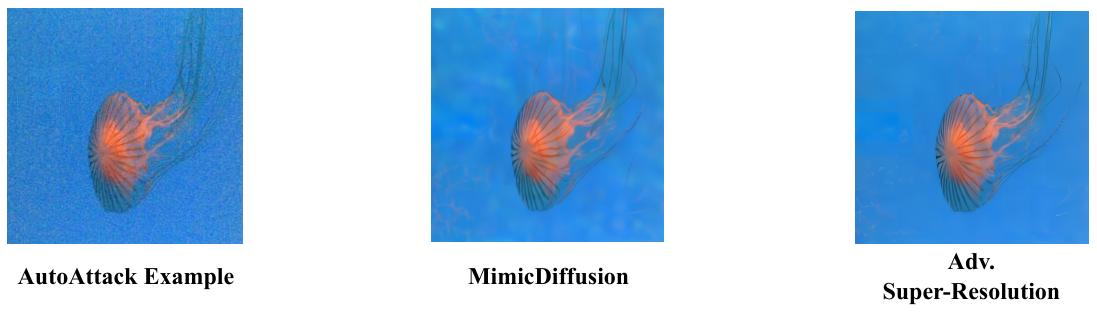}
      \caption{\textbf{The example of proposed adversarial super-resolution.} Our method achieves similar adversarial purification without any gradient calculation of diffusion models. It is worth noting that our adversarial super-resolution achieves superior restoration of background regions in the image.}
   \label{fig:3}
    \vspace{-2mm}
   \end{figure*}

During the adversarial anti-aliasing phase, we significantly reduce adversarial perturbations by directly decreasing pixel-wise modifications of adversarial examples. However, this approach may not be effective against unrestricted adversarial examples, as they are not generated by adding explicit perturbations. Additionally, blurring the images can negatively impact the clean accuracy of the target model. Super-resolution offers an effective way to recover high-quality images from our adversarial anti-aliased images. Previous super-resolution methods \cite{ledig2017photo,gao2019multi} typically modify original pixels of the low-resolution image and use the residual features of the original low-resolution image. These methods can inadvertently transfer negative effects from the adversarial examples to the final high-resolution images, making them ineffective for adversarial super-resolution. Diffusion-based super-resolution \cite{yue2024resshift,rombach2022high} provides a more isolated approach for super-resolution. These models generate high-resolution images through a denoising-like process over randomly sampled noise, using the low-resolution image as conditions. 

In this work, we adopt the ResShift method by Yue et al. \cite{yue2024resshift} for our super-resolution process. This super-resolution model can incorporate benign priors for defense, as it is trained with the clean dataset of the target model. ResShift is a diffusion-based model for image restoration, designed to learn the residual transformation $\bm{e}_0 = \bm{x}_0 - \bm{X}$ that maps a low-resolution (LR) image $\bm{x}_0$ to its corresponding high-resolution (HR) image $\bm{X}$. The residual $\bm{e}_0$ is scheduled over time steps ${\eta_t}$, where $t = 1, 2, \dots, T$ denotes the diffusion timesteps. The forward process of ResShift is defined as:
\begin{equation}
    q(\bm{x}_t|\bm{x}_{t-1},\bm{x}_0) = \mathcal{N}(\bm{x}_t; \bm{x}_{t-1}+\alpha_t \bm{e}_0, \kappa^2 \alpha_t \textbf{I})
    \label{eq:transit_t_t1}
\end{equation}
where $\alpha_t=\eta_t-\eta_{t-1}$ for $t>1$ and $\alpha_1=\eta_1$, and $\kappa$ is a hyperparameter.

The forward process is formulated as follows:
\begin{equation}
    p_\theta(\bm{x}_{t-1}|\bm{x}_t, \bm{x}_0) = \mathcal{N}(\bm{x}_{t-1}:\bm{\mu}_{\theta}(\bm{x}_t,\bm{x}_0,t),\bm{\Sigma}_{\theta}(\bm{x}_t,\bm{x}_0,t)) 
\end{equation}
where $\bm{\mu}_{\theta}$ and $\bm{\Sigma}$ are calculated with the trained deep learning model $\epsilon_{\bm{\theta}}$:
\begin{equation}
    \bm{\mu}_{\bm{\theta}}(\bm{x}_t,\bm{x}_0,t) = \frac{\eta_{t-1}}{\eta_t}\bm{x}_t+\frac{\alpha_t}{\eta_t}\epsilon_{\bm{\theta}}(\bm{x}_t,\bm{x}_0,t),
    \label{eq:repqrameter_mean_reparameter}
\end{equation}
\begin{equation}
\bm{\Sigma}_{\bm{\theta}}(\bm{x}_t,\bm{x}_0,t)=\kappa^2\frac{\eta_{t-1}}{\eta_t}\alpha_t\textbf{I}
\end{equation}

Figure \ref{fig:3} demonstrates that the proposed super-resolution method achieves results comparable to diffusion-based adversarial purification \cite{song2024mimicdiffusion}, which do not require the calculation of gradient. 



\subsubsection{Adversarial Purification}

The proposed adversarial purification is performed by the combination of adversarial anti-aliasing and adversarial super-resolution. We resize the purified images after the adversarial super-resolution for the target model. Additionally, our approach does not require any training of the target model or the defense model.

\begin{equation}
    y=\{f(\text{SR}( \text{AA}(\bm{x}_\text{adv})) ))\}
\end{equation}

\subsection{Comparisons to DiffPure}

From the perspective of diffusion models, our adversarial super-resolution approach is similar to DiffPure \cite{nie2022DiffPure}, as both utilize backward processes of diffusion models for purification starting from adversarial images as input. However, DiffPure does not employ the entire diffusion process for purification because the diffusion model of DiffPure focuses on the image generation task and begins with Gaussian noise, which can result in the loss of label semantics when using the full diffusion process. Consequently, their defense performance is inferior to methods that use the complete diffusion process, such as MimicDiffusion \cite{song2024mimicdiffusion}.

Our adversarial super-resolution can leverage the entire diffusion process because it starts with a low-resolution image. This approach overcomes the limitations of DiffPure, which uses only a partial diffusion process. Additionally, super-resolution tasks ensure that the purified image remains consistent with the input image by relying on its low-resolution input, rather than using gradient guidance. This method is more computationally efficient than guidance-based purification, such as MimicDiffusion.

Another reason super-resolution yields better performance is due to the training process. The super-resolution diffusion model in our study is trained using low-resolution images generated by blurring with a filter and adding Gaussian noise. The blurring with a filter helps learn prior knowledge from clean data, similar to our proposed adversarial anti-aliasing. The addition of Gaussian noise can be seen as introducing adversarial perturbations to the clean images. This prior knowledge enables super-resolution to be effectively used for adversarial purification. Furthermore, it suggests that training the diffusion model with adversarial priors can enhance the performance of purification. 

We give a trivial theoretical analysis of why adversarial super-resolution tasks achieve better performance than DiffPure's purification. In DiffPure's paper, it is proved that for clean distribution $\bm{x}_t \sim p(x)$ and adversarial distribution $\bm{\hat{x}}_t \sim q(x)$ at diffusion timestep $t$, we have:
\begin{equation}
    \frac{\partial D_{KL}(\bm{x}_t||\bm{\hat{x}}_t ) }{\partial t } \le 0
\end{equation}
where the equality happens only when $x_t$ and $\hat{x}_t$ are sampled from the same distribution, i.e., from the initial timestep $T$. In other words, the difference in the distribution of $p(x)$ and $q(x)$ increases during the reverse diffusion process.

Therefore, it is trivial that utilizing the whole diffusion process from timestep $T$ achieves better performance to recover $\bm{x}_0$ from $p(\bm{x})$ than starting from an intermediate timestep $t^*$, where:
\begin{equation}
     D_{KL}(\bm{x}_T||\bm{\hat{x}}_T)< D_{KL}(\bm{x}_t^*||\bm{\hat{x}}_t^*)
     \label{eq:a1}
\end{equation}

Equation \ref{eq:a1} demonstrates that our adversarial purification method is superior to DiffPure, which starts from $\hat{x}_T$ for purification while DiffPure relies on intermediate timestep $\hat{x}_t^*$ for purification. However, as ResShift uses only 15 steps for super-resolution, it is important to use adversarial anti-aliasing before the adversarial super-resolution for the reason ResShift achieves fewer diffusion-based purification steps. Compared to guidance-based purification like MimicDiffusion, which also uses the whole diffusion timestep for purification, our method is significantly time-efficient without using a gradient of the diffusion model to maintain the consistency between the purified image and the adversarial image.

\section{Gradient-Free Adversarial Deblurring Fine-Tuning} 
\label{Debl}
In Section~\ref{Puri}, we employ the pre-trained ResShift checkpoint for adversarial purification. Intuitively, the performance of ResShift could be further improved by fine-tuning it on the adversarial deblurring task using the target dataset, whereas the original ResShift model is trained solely on the ImageNet dataset.

\subsection{Deblurring Fine-tuning}

Unlike the super-resolution task, which reconstructs a high-resolution image from a sub-sampled low-resolution input, the deblurring task is more suitable for adversarial purification, as it aims to recover the original image from its blurred counterpart while preserving the same resolution. For fine-tuning, we adopt our adversarial anti-aliasing operator as the deblurring kernel. The fine-tuning loss is defined in accordance with ResShift:
\begin{equation}
    \mathcal{L}(\bm{x}_t,\bm{x}_0,t)=\sum\nolimits_t\Vert \hat{\bm{x}}^t_0 - \bm{X} \Vert_2^2 + \lambda l_p(\hat{\bm{x}}^t_0 - \bm{X}), \bm{x}_0=\text{AA}(\bm{X})
    \label{eq:ft}
\end{equation}
where $\hat{\bm{x}}^t_0$ is the prediction of $\bm{x}_0$ at timestep $t$ by $\epsilon_\theta$ and $l_p$ is the LPIPS metric.

\subsection{Contrastive Learning-based Adversarial Deblurring Fine-tuning}
To integrate the adversarial purification objective, we incorporate contrastive learning into the deblurring fine-tuning process. Recent studies have shown that contrastive learning achieves superior performance in unsupervised representation learning tasks by pulling representations of positive pairs closer while pushing those of negative pairs farther apart. In this work, we adopt the InfoNCE \cite{chen2020simple} contrastive objective for fine-tuning. Given a blurred image $\bm{x}_0 = \text{AA}(\bm{X})$, we construct the positive sample $\bm{x}_0^+$ as $\bm{x}_0^+=\text{AA}(\bm{X}+\bm{\varepsilon}),\bm{\varepsilon} \sim \mathcal{N}(0,\mathbf{I})$ where the positive sample is generated by adding Gaussian noise to the high-resolution image $\bm{X}$. For negative pairs $\bm{x}_0^-$, we employ adversarial examples generated by PGD and AutoAttack with $\ell_\infty = 4/255$ and $\ell_2 = 4/255$, thereby incorporating adversarial deblurring into the fine-tuning task. The final learning objective is formulated as:

\begin{equation}
    \mathcal{L}_\text{CL}=-\log\frac{\exp(\text{sim}(\bm{x}_t,\bm{x}_t^+)/\tau)}{\sum_{n \in(\bm{x}_t^+,\bm{x}_t^-)}\exp(\text{sim}(\bm{x}_t,\bm{x}_t^n)/\tau)}
    \label{eq:cl}
\end{equation}
where $\tau$ is the temperature parameter, and $\text{sim}$ is the cosine similarity from the $\epsilon_\theta$ in feature space:
\begin{equation}
    \text{sim}(\bm{x},\bm{y})=\frac{\epsilon_\theta(\bm{x})\cdot\epsilon_\theta(\bm{y})}{\Vert\epsilon_\theta(\bm{x})\Vert - \Vert\epsilon_\theta(\bm{y}) \Vert}
\end{equation}

The detailed fine-tuning procedure is summarized in Algorithm~\ref{alg:ft}. To mitigate overfitting, we inject Gaussian noise into the training data.

\begin{algorithm}[t]
    \caption{{Adversarial Deblurring Fine-tuning}}
    \label{alg:ft}
    \begin{algorithmic}[1]
        \Require {Target dataset $\mathcal{D}$, Pre-trained model $\epsilon_\theta$, Pre-trained ResNet-50 model $f(\cdot)$}
        \Repeat
        \State {$\bm{X} \sim \mathcal{T}$, $\bm{x}_0 = \text{AA}(\bm{X}+\bm{\varepsilon}),~\bm{\varepsilon} \sim \mathcal{N}(0,\mathbf{I})$}
        \State Construct positive sample
        \State $\bm{x}_0^+=\text{AA}(\bm{X}+\bm{\varepsilon}),\bm{\varepsilon} \sim \mathcal{N}(0,\mathbf{I})$
        \State Construct negative sample against $f(\cdot)$
        \State $\bm{x}_0^-=\text{AA}(\text{Adv}(\bm{X})),\text{Adv}\in{\{\text{PGD},\text{AutoAttack}\}},|\text{Adv}|=\{\ell_\infty:4/255, \ell_\text{2}:4/255\}$
        \State {$t \sim \text{Uniform}\left(\{1,\cdots,T\}\right)$}
        \State {$\bm{x}_t \sim q(\bm{x}_t | \bm{x}_0, \bm{y}_0)$}
        \State Calculate $\mathcal{L}$ with Equation \ref{eq:ft}
        \State {$\bm{x}_t^+ \sim q(\bm{x}_t | \bm{x}_0, \bm{x}_0^+)$},{$\bm{x}_t^- \sim q(\bm{x}_t | \bm{x}_0, \bm{x}_0^-)$}
        \State Calculate $\mathcal{L}_{\text{CL}}$ with Equation \ref{eq:cl}
        \State Fine-tuning $\epsilon_\theta$ with $\nabla ({\mathcal{L}+}\mathcal{L}_{\text{CL}})$
        \Until{{converged}}
    \end{algorithmic}
\end{algorithm}

\section{Experiments}

\subsection{Experimental Setup}

\begin{table*}[t]
\caption{\textbf{The standard and robust accuracy against left: AutoAttack ($\ell_\infty=8/255$), right: PGD+EOT ($\ell_\infty=8/255$) on the CIFAR-10 dataset. } The PGD+EOT is performed in white-box attack scenario with the surrogate process. }

\label{tab:cifaraa}
\centering
\resizebox{1.5\columnwidth}{!}{  
\begin{tabular}{l c c c c}
        \toprule
         Method & Target Model & Standard Acc(\%) & AutoAttack Acc(\%) & PGD+EOT Acc(\%) \\
        \midrule
        Wu \textit{et al.}~\cite{wu2020adversarial} & WideResNet-28-10 & 85.36 & 59.18 & 62.16 \\
        Gowal \textit{et al.}~\cite{gowal2021improving} & WideResNet-28-10 & 87.33 & 61.72 & 64.68\\
        Rebuffi \textit{et al.}~\cite{rebuffi2021data} & WideResNet-28-10 & 87.50 & 65.24 & 68.89 \\
        Wang \textit{et al.}~\cite{wang2022guided} & WideResNet-28-10 & 84.85 & 71.18 & 68.36 \\
        Nie \textit{et al.}~\cite{nie2022DiffPure} & WideResNet-28-10 & 89.23 & 71.03 & 46.84\\
        Lee \textit{et al.}~\cite{lee2023robust} & WideResNet-28-10 & 90.16 & 70.47 & 55.82 \\
        Chen \textit{et al.}~\cite{chen2024diffilter} & WideResNet-28-10 & 91.46 & 64.06 & 57.83 \\
        Song \textit{et al.}~\cite{song2024mimicdiffusion} & WideResNet-28-10 &  92.10 & 75.45 & 68.20 \\\midrule
        Ours & WideResNet-28-10 & 92.54 $\pm$ 1.66 &   82.02  $\pm$  1.17 & 80.86  $\pm$  1.33\\
        Ours$_\text{DB}$ & WideResNet-28-10 & \textbf{94.31} $\pm$ \textbf{1.41} &   \textbf{85.57}  $\pm$  \textbf{1.04} & \textbf{81.23}  $\pm$  \textbf{0.89}\\
        \midrule
        Rebuffi \textit{et al.}~\cite{rebuffi2021data} & WideResNet-70-16 & 88.54 & 64.46 & 68.23 \\
        Gowal \textit{et al.}~\cite{gowal2021improving}& WideResNet-70-16 & 88.74 &  66.60 & 69.48 \\
        Nie \textit{et al.}~\cite{nie2022DiffPure} & WideResNet-70-16 & 91.04 & 71.84 & 51.13 \\
        Lee \textit{et al.}~\cite{lee2023robust} & WideResNet-70-16 & 90.43 & 66.06 & 56.88 \\
        Bai \textit{et al.}~\cite{baidiffusion} & WideResNet-70-16 & 92.97 & 82.81 & 73.21 \\
        Song \textit{et al.}~\cite{song2024mimicdiffusion} & WideResNet-70-16 & 93.25 &  76.60 & 69.55\\\midrule
        Ours & WideResNet-70-16 & 93.42 $\pm$ 1.51 &   83.65  $\pm$ 2.90 &   81.60 $\pm$  1.75 \\
        Ours$_\text{DB}$ & WideResNet-70-16 & \textbf{94.40} $\pm$ \textbf{1.14} &   \textbf{86.53}  $\pm$  \textbf{1.89} &   \textbf{82.58}  $\pm$  \textbf{1.20} \\
        \bottomrule
    \end{tabular}
}
\vspace{-2mm}
\end{table*}

\begin{table}[t]
\caption{\textbf{The defense performance against AutoAttack ($\ell_\text{2}=0.5$) on the CIFAR-10 dataset.} }
\label{tab:l2cifar}
\centering
\resizebox{1.0\columnwidth}{!}{  
\begin{tabular}{l c c c}
        \toprule
         Method & Target Model & Standard Accuracy(\%) & Robust Accuracy(\%) \\
        \midrule
        Rony \textit{et al.}~\cite{rony2019decoupling} & WideResNet-28-10 & 89.05 & 66.41 \\
        Ding \textit{et al.}~\cite{ding2020mma} & WideResNet-28-10 & 88.02 &  67.77 \\
        Rebuffi \textit{et al.}~\cite{rebuffi2021data} & WideResNet-28-10 & 87.50 & 78.32 \\
        Nie \textit{et al.}~\cite{nie2022DiffPure} & WideResNet-28-10 & 89.23 & 78.98 \\
        Song \textit{et al.}~\cite{song2024mimicdiffusion} & WideResNet-28-10 & 92.10 &  81.52 \\\midrule
        Ours & WideResNet-28-10 & 92.54 $\pm$ 1.66 &   84.90  $\pm$  2.82 \\
        Ours$_\text{DB}$ & WideResNet-28-10 & \textbf{94.31} $\pm$ \textbf{1.41} &   \textbf{86.19}  $\pm$  \textbf{2.37} \\
        \midrule
        Rebuffi \textit{et al.}~\cite{rebuffi2021data} & WideResNet-70-16 & 88.54 & 80.86\\
        Gowal \textit{et al.}~\cite{gowal2021improving} & WideResNet-70-16 & 88.74 & 74.03\\
        Nie \textit{et al.}~\cite{nie2022DiffPure} & WideResNet-70-16 & 91.04 & 81.17\\
        Song \textit{et al.}~\cite{song2024mimicdiffusion} & WideResNet-70-16 & 93.25 & 83.60  \\ \midrule
        Ours & WideResNet-70-16 & 93.42 $\pm$ 1.51 &   87.60  $\pm$  2.35 \\
        Ours$_\text{DB}$ & WideResNet-70-16 & \textbf{94.40} $\pm$ \textbf{1.14} &   \textbf{88.14}  $\pm$  \textbf{2.02} \\
        \bottomrule
    \end{tabular}
}
\vspace{-2mm}    
\end{table}

\begin{table}[t]
\caption{\textbf{The defense performance against C\&W ($\ell_\text{2}=8/255$, EOT=50) on the CIFAR-10 dataset.} }

\label{tab:cwcifar}
\centering
\resizebox{1.0\columnwidth}{!}{  
\begin{tabular}{l c c c}
        \toprule
         Method & Target Model & Standard Accuracy(\%) & Robust Accuracy(\%) \\
        \midrule
        Nie \textit{et al.}~\cite{nie2022DiffPure} & WideResNet-28-10 & 89.23 & 47.65 \\
        Wang \textit{et al.}~\cite{wang2022guided} & WideResNet-28-10 & 84.85 & 21.71 \\
        Song \textit{et al.}~\cite{song2024mimicdiffusion} & WideResNet-28-10 & 92.10 &  89.91 \\\midrule
        Ours & WideResNet-28-10 & 92.54 $\pm$ 1.66 &   90.51  $\pm$  4.52 \\
        Ours$_\text{DB}$ & WideResNet-28-10 & \textbf{94.31} $\pm$ \textbf{1.41} &   \textbf{91.30}  $\pm$  \textbf{3.88} \\
        \bottomrule
    \end{tabular}
}

\vspace{-2mm}
\end{table}

\begin{table}[t]
\caption{\textbf{The standard and robust accuracy against BPDA ($\ell_\infty=8/255$) on the CIFAR-10 dataset with WideResNet-28-10 as the target model. } }

\centering
\resizebox{1.0\columnwidth}{!}{  
\begin{tabular}{l c c c}
        \toprule
         Method & Purification & Standard Acc(\%) & Robust Acc(\%) \\
        \midrule
        Nie \textit{et al.}~\cite{nie2022DiffPure}($t^{\ast} = 0.0075$) & Diffusion & 91.38 & 77.62 \\
        Nie \textit{et al.}~\cite{nie2022DiffPure}($t^{\ast} = 0.1$) & Diffusion & 89.23 & 81.56 \\
        Wang \textit{et al.}~\cite{wang2022guided} & Diffusion & 84.45 & 77.31\\
        Chen \textit{et al.}~\cite{chen2024diffilter} & Diffusion & 91.46 & 64.73  \\
        Song \textit{et al.}~\cite{song2024mimicdiffusion} & Diffusion & 92.10 & 76.45 \\\midrule
        Ours & Diffusion & 92.54 $\pm$ 1.66 &   81.24  $\pm$  2.51 \\
        Ours$_\text{DB}$ & Diffusion & \textbf{94.31} $\pm$ \textbf{1.41} &   \textbf{83.45}  $\pm$  \textbf{2.44} \\
        \bottomrule
    \end{tabular}
}
   
\label{tab:cifareot}
\vspace{-2mm}
\end{table}

\begin{table}[t]
\caption{\textbf{The standard and robust accuracy against AdvDiff on the CIFAR-10 dataset. } }

 \centering
\resizebox{0.99\columnwidth}{!}{  
\begin{tabular}{l c c c}
        \toprule
         Method & Target Model & Standard Acc(\%) & Robust Acc(\%) \\
        \midrule
        Nie \textit{et al.}~\cite{nie2022DiffPure} & WideResNet-28-10 & 95.42 & 21.56 \\
        Wang \textit{et al.}~\cite{wang2022guided} & WideResNet-28-10 & 95.86 & 26.68 \\
        Lee \textit{et al.}~\cite{lee2023robust} & WideResNet-28-10 & 95.29 & 24.94 \\
        Song \textit{et al.}~\cite{song2024mimicdiffusion} & WideResNet-28-10 & 96.21 &  23.23  \\ \midrule
        Ours & WideResNet-28-10 & 96.80 $\pm$ 0.37 & 33.97 $\pm$ 0.77 \\
        Ours$_\text{DB}$ & WideResNet-28-10 & \textbf{96.92} $\pm$ \textbf{0.28} &  \textbf{37.66}  $\pm$ \textbf{0.54} \\
        \bottomrule
    \end{tabular}
}
   
\label{tab:cifaruae}
\vspace{-2mm}
\end{table}

\noindent \textbf{Dataset and target models.} We consider CIFAR-10 \cite{krizhevsky2009learning} and ImageNet \cite{deng2009imagenet} for major evaluation. The CIFAR-10 dataset consists of 32 × 32 color images across 10 classes, with 50,000 training samples and 10,000 test samples. The ImageNet subset contains 256 × 256 color images from 1000 classes with 1.2 million data in total. For target models, we adopt WideResNet-28-10 and
WideResNet-70-16 \cite{zagoruyko2016wide} for CIFAR-10 dataset and ResNet50 \cite{he2016deep} for ImageNet dataset. Vision transformers are further included in the experiment on ImageNet dataset. These are commonly adopted backbones for adversarial robustness evaluation.

\noindent \textbf{Comparisons.} We compared our defense methods with various state-of-the-art defenses by the standardized benchmark: RobustBench \cite{croce2021robustbench}. We compare four diffusion-based adversarial purification methods: Nie et al.'s DiffPure \cite{nie2022DiffPure}, Wang et al.'s \cite{wang2022guided}, Lee et al.'s \cite{lee2023robust} and Song et al.'s MimicDiffusion \cite{song2024mimicdiffusion}. We mainly compare our method with MimicDiffusion as it is the current state-of-the-art method. We use the Score SDE \cite{songscore} implementation of MimicDiffusion on CIFAR-10 for fair comparisons. The defense methods that use extra data are not compared for fairness. We only evaluate the adversarial purification methods against unrestricted adversarial attacks as the adversarial training's different threat model. 

\noindent \textbf{Attack settings.} We evaluate our method with both perturbation-based attacks and diffusion-based unrestricted adversarial attacks. For perturbation-based attacks, we select AutoAttack \cite{croce2020reliable} and PGD \cite{madry2017towards}. For diffusion-based unrestricted adversarial attacks, we use DiffAttack \cite{chen2023diffusion} and AdvDiff \cite{dai2023advdiff} for comparisons. DiffAttack is only evaluated on the ImageNet dataset according to the original paper. To ensure a fair comparison with previous diffusion-based adversarial purification, we include the evaluation against the adaptive attack, i.e., reverse pass differentiable approximation (BPDA) \cite{hillstochastic}. We also evaluate the performance against PGD+EOT that is discussed in \cite{lee2023robust}. We evaluate adaptive attacks in a white-box setting, where the attacker has access to the gradient of the purification model. However, since our defense employs in-place operations that clip gradients in the adversarial anti-aliasing, we restrict gradient computation to the diffusion-based purification component. On CIFAR-10, the attack settings follow DiffPure \cite{nie2022DiffPure}. On ImageNet, we randomly sample 5 images from each class and average over 10 runs. The PGD+EOT settings all follow Lee et al. \cite{lee2023robust} with the surrogate process.

\noindent \textbf{Implementation details.} We adopt the mean filter with $[[1,1],[1,1]]$ for adversarial anti-aliasing on CIFAR-10, and $[[1,1,1,1,1],[1,1,0,1,1],[1,1,1,1,1]]$ in ImageNet. ResShift \cite{yue2024resshift} is utilized for adversarial super-resolution, and the official ImageNet v2 parameter checkpoint is used for both ImageNet and CIFAR-10 defenses. Our experiment is implemented with PyTorch on two NVIDIA GeForce RTX 3090 GPUs. For the CIFAR-10 dataset, we upscale the adversarial anti-aliased images with PyToch to $64 \times 64$ resolution for ResShift. For deblurring fine-tuning, we select the training set from CIFAR-10 and ImageNet dataset to construct the dataset. We use Ours$_\text{DB}$ to denote the propose defense with deblurring fine-tuning.
We use the official Score SDE \cite{songscore} checkpoint for CIFAR-10 and LDM \cite{rombach2022high} checkpoint for ImageNet to generate UAEs.

\noindent \textbf{Evaluation metrics.} Following Nie et al. \cite{nie2022DiffPure}, we use \textit{standard accuracy} and \textit{robust accuracy} as the evaluation metrics, where \textit{standard accuracy} measures the clean accuracy against the benign data after the purification and \textit{robust accuracy} measures the clean accuracy against the  adversarial examples after the purification. Both are calculated according to the top-1 classification accuracy. The adversarial examples are generated from the test set.

\subsection{Attack Performance}
\subsubsection{CIFAR-10}
\noindent \textbf{Perturbation-based adversarial attack}. Table \ref{tab:cifaraa}, \ref{tab:l2cifar}, \ref{tab:cwcifar} presents the defense performance against AutoAttack, PGD, and C\&W attacks on the CIFAR-10 dataset. The results demonstrate that our proposed method achieves better standard accuracy and robust accuracy than previous attack methods with our baseline defense. By fine-tuning over the CIFAR-10 dataset, our defense can further improve the defense performance, which validate the effectiveness of the proposed fine-tuning approach. Because images in the CIFAR-10 dataset are only with $32 \times 32$ resolution, we set our anti-aliasing filter to a relatively small size. Table \ref{tab:cifareot} indicates that the robustness performance of the proposed method is on par with the state-of-the-art method \cite{nie2022DiffPure}. This finding suggests that our method is more suitable for high-resolution images, as $32 \times 32$ may not be large enough to effectively extract the semantic shape for our approach.  However, we can further enhance our performance by incorporating adversarial purification techniques from previous work.

Our defense's performance against PGD+EOT and BDPA on Table \ref{tab:cifaraa} and \ref{tab:cifareot} showcases its ability to defend against adaptive attacks. This is because our approach focuses on extracting and recovering the semantic features from adversarial images, rather than inferring and denoising the adversarial perturbations. Moreover, the proposed adversarial anti-aliasing employs in-place operations to block gradient propagation to the attacker, thereby significantly enhancing the robust accuracy of diffusion-based purification under white-box attack scenarios. Consequently, our defense achieves comparable effectiveness against both adaptive and standard attacks. We further suggest that existing diffusion-based purification methods can incorporate our pre-processing approach to better maintain robustness against adaptive attacks.

\begin{table*}[t]
\caption{\textbf{The standard and robust accuracy against left: PGD ($\ell_\infty=4/255$), right: PGD+EOT ($\ell_\infty=4/255$) on the ImageNet dataset.} The PGD+EOT is performed in white-box attack scenario with the surrogate process. }

 \centering
\resizebox{1.4\columnwidth}{!}{  
\begin{tabular}{l c c c c}
        \toprule
         Method & Target Model & Standard Acc(\%) & PGD Acc(\%)  & PGD+EOT Acc(\%)  \\
        \midrule
        Wong \textit{et al.}~\cite{wong2020fast} & ResNet50 & 55.62 & 26.24 & 30.51 \\
        Salman \textit{et al.}~\cite{salman2020do} & ResNet50 & 64.02 &  34.96 & 38.62 \\
        Bai \textit{et al.}~\cite{bai2021recent} & ResNet50 & 67.38 & 40.27 & 43.42 \\
        Nie \textit{et al.}~\cite{nie2022DiffPure} & ResNet50 & 68.22 & 42.88 & 38.71 \\
        Lee \textit{et al.}~\cite{lee2023robust} & ResNet50 & 70.74 & 46.31 & 42.15 \\
        Wang \textit{et al.}~\cite{wang2022guided} & ResNet50 & 70.17 & 68.78 & 40.22 \\
        Chen \textit{et al.}~\cite{chen2024diffilter} & ResNet50 & 74.24 & \textbf{71.64} & 53.04 \\
        Song \textit{et al.}~\cite{song2024mimicdiffusion} & ResNet50 & 66.92 & 62.16 & 52.66 \\\midrule
        Ours & ResNet50 & \textbf{75.28} $\pm$ \textbf{1.06} &  69.75  $\pm$ 2.61 &  66.87  $\pm$ 1.85 \\
        Ours$_\text{DB}$ & ResNet50 & 74.89 $\pm$ 1.10 &  71.10  $\pm$ 2.61  &  \textbf{68.13}  $\pm$ \textbf{1.79} \\
        \bottomrule
    \end{tabular}
}
   
\label{tab:imagenetpgd}
\vspace{-2mm}
\end{table*}

\noindent \textbf{Unrestricted adversarial attack}. 
Unrestricted adversarial examples on the CIFAR-10 dataset are challenging to defend against, as shown in Table \ref{tab:cifaruae}. Our purification method outperforms the previous adversarial purification approach \cite{song2024mimicdiffusion} by an average of 10\%, validating the effectiveness of our proposed defense. Unrestricted adversarial examples are generated using the same diffusion models employed for purification, with adversarial guidance injected into the sampling process. Since diffusion-based purification relies on a benign generation process that overlooks adversarial guidance, such perturbations are difficult to remove. In contrast, our proposed defense leverages super-resolution diffusion models with a distinct diffusion process. More effectively, the adversarial anti-aliasing module further mitigates adversarial guidance through its blurring-based purification.

\subsubsection{ImageNet}

\begin{table}[t]
\caption{\textbf{The standard and robust accuracy against AutoAttack ($\ell_\infty=8/255$) on the ImageNet dataset.} }

 \centering
\resizebox{1.0\columnwidth}{!}{  
\begin{tabular}{l c c c}
        \toprule
         Method & Target Model & Standard Acc(\%) & Robust Acc(\%) \\
        \midrule
        Engstrom \textit{et al.}~\cite{croce2021robustbench} & ResNet50 & 62.56 & 31.06  \\
        Wong \textit{et al.}~\cite{wong2020fast} & ResNet50 & 55.62 & 26.95 \\
        Salman \textit{et al.}~\cite{salman2020do} & ResNet50 & 64.02 &  37.89 \\
        Bai \textit{et al.}~\cite{bai2021recent} & ResNet50 & 67.38 & 35.51 \\
        Nie \textit{et al.}~\cite{nie2022DiffPure} & ResNet50 & 68.22 & 43.89  \\
        Bai \textit{et al.}~\cite{baidiffusion}  & ResNet50 & 70.41 & 45.59\\
        Chen \textit{et al.}~\cite{chen2024diffilter} & ResNet50 & 74.24 & 43.39 \\
        Song \textit{et al.}~\cite{song2024mimicdiffusion} & ResNet50 & 66.92 &  61.53 \\ \midrule
        Ours & ResNet50 & \textbf{75.28} $\pm$ \textbf{1.06} & 67.61  $\pm$ 1.95 \\
        Ours$_\text{DB}$ & ResNet50 & 74.89 $\pm$ 1.10 & \textbf{68.80}  $\pm$ \textbf{1.84} \\
        
        \bottomrule
    \end{tabular}
}
   
\label{tab:imagenetaa}
\vspace{-2mm}
\end{table}

\begin{table}[t]
\caption{\textbf{The defense performance (Robust Accuracy) against AutoAttack ($\ell_\infty=8/255$) on the ImageNet dataset against vision transformers.} }

 \centering
\resizebox{1.0\columnwidth}{!}{  
\begin{tabular}{lcccc}

\toprule

Method             &  Vit-B(\%)  & Deit-B(\%) & Swin-B(\%) & Average(\%)  \\ \midrule
Nie \textit{et al.}~\cite{nie2022DiffPure}      & 32.15      & 28.32    &  27.48 & 29.32 \\ 
Lee \textit{et al.}~\cite{lee2023robust}      & 35.92      & 31.34    &  30.87 & 32.71 \\ 
Wang \textit{et al.}~\cite{wang2022guided}      & 42.71      & 36.50    &  37.48  & 38.90 \\ 
Song \textit{et al.}~\cite{song2024mimicdiffusion}  & 44.30    & 35.41 & 38.52  & 39.41 \\ \midrule
Ours  & \textbf{68.42}    & \textbf{57.84} & \textbf{60.77}  & \textbf{62.34} \\
\bottomrule
\end{tabular}

}
   
\label{tab:vit}
\vspace{-2mm}
\end{table}

\noindent \textbf{Perturbation-based adversarial attack}. ImageNet serves as a more realistic and challenging benchmark for evaluating purification performance, as it consists of 1,000 diverse object categories and high-resolution image data that better reflect real-world scenarios. As reported in Tables \ref{tab:imagenetaa} and \ref{tab:imagenetpgd}, our proposed defense achieves a substantial improvement in both standard accuracy and robust accuracy over existing methods. Notably, the standard accuracy of our defense clearly surpasses prior approaches, demonstrating that adversarial super-resolution effectively leverages prior knowledge learned from the training distribution to enhance classification performance under clean conditions.
Beyond clean accuracy, the proposed adversarial anti-aliasing mechanism proves particularly effective on the ImageNet dataset. By applying filtering operations, it suppresses adversarial perturbations that are often hidden in the fine-grained pixel structures of high-resolution adversarial examples. This highlights the importance of incorporating frequency-aware defenses when dealing with datasets characterized by detailed textures and complex spatial patterns. In addition, our deblurring-based fine-tuning method further improves robustness against perturbation-based attacks, confirming its general effectiveness across diverse adversarial threat models.

Table \ref{tab:vit} further illustrates the strength of our defense when applied to vision transformers. Compared with state-of-the-art diffusion-based purification methods, our approach consistently achieves substantially higher accuracy, underscoring its adaptability to different backbone architectures and its ability to scale effectively with modern large-capacity models.

Finally, the results against adaptive PGD+EOT attacks provide strong evidence of the overall robustness of our pipeline. While such adaptive attacks are theoretically powerful, their execution requires prohibitive computational overhead in terms of both runtime and memory, rendering them impractical for real-time adversarial scenarios.

\begin{table}[t]
\caption{\textbf{The standard and robust accuracy against AdvDiff on the ImageNet dataset.} }

 \centering
\resizebox{0.99\columnwidth}{!}{  
\begin{tabular}{l c c c}
        \toprule
         Method & Target Model & Standard Acc(\%) & Robust Acc(\%) \\
        \midrule
        Nie \textit{et al.}~\cite{nie2022DiffPure} & ResNet50 & 91.48 & 24.82  \\
        Wang \textit{et al.}~\cite{wang2022guided} & ResNet50 & 92.31 & 26.74 \\
        Lee \textit{et al.}~\cite{lee2023robust} & ResNet50 & 91.80 & 25.34  \\
        Song \textit{et al.}~\cite{song2024mimicdiffusion} & ResNet50 & 92.54 &  25.35 \\\midrule
        Ours & ResNet50 & \textbf{97.83} $\pm$ \textbf{1.36} &  42.21  $\pm$ 3.41 \\
        Ours$_\text{DB}$ & ResNet50 & 97.55 $\pm$ 1.28 & \textbf{43.09}  $\pm$ \textbf{3.19} \\
        \bottomrule
    \end{tabular}
}
\label{tab:imagenetuae}
\vspace{-2mm}
\end{table}

\begin{table}[t]
\caption{\textbf{The standard and robust accuracy against DiffAttack on the ImageNet dataset.} }

 \centering
\resizebox{0.99\columnwidth}{!}{  
\begin{tabular}{l c c c}
        \toprule
         Method & Target Model & Standard Acc(\%) & Robust Acc(\%) \\
        \midrule
        Nie \textit{et al.}~\cite{nie2022DiffPure} & ResNet50 &  68.22 & 59.15  \\
        Wang \textit{et al.}~\cite{wang2022guided} & ResNet50 & 69.54 & 62.33 \\
        Lee \textit{et al.}~\cite{lee2023robust} & ResNet50 & 70.74 & 61.56  \\
        Song \textit{et al.}~\cite{song2024mimicdiffusion} & ResNet50 & 66.92 &  60.17\\\midrule
        Ours & ResNet50 & \textbf{75.28} $\pm$ \textbf{1.06} & 65.51  $\pm$ 1.33  \\
        Ours$_\text{DB}$ & ResNet50 & 74.89 $\pm$ 1.10 &  \textbf{66.37}  $\pm$ \textbf{1.20}  \\
        \bottomrule
    \end{tabular}
}
   
\label{tab:imagenetuaed}
\vspace{-2mm}
\end{table}

\noindent \textbf{Unrestricted adversarial attack}. We evaluate the defense performance of various methods against the unrestricted adversarial attacks AdvDiff and DiffAttack, with results reported in Tables \ref{tab:imagenetuae} and \ref{tab:imagenetuaed}. The findings demonstrate that existing defenses remain largely ineffective against these recently proposed unrestricted adversarial attacks.
For AdvDiff (Table \ref{tab:imagenetuae}), the standard accuracy is measured using images generated by the benign sampling process of the latent diffusion model (LDM). The relatively high standard accuracy observed in this setting is primarily due to the strong generative capability of diffusion models, which are trained on the training set of ImageNet dataset and thus inherently produce samples closely aligned with the training distribution. This alignment enables target classifiers to achieve performance comparable to their training accuracy. Importantly, our defense not only preserves this high standard accuracy but also achieves substantially higher robust accuracy than prior defenses. This improvement arises from leveraging a distinct prior different from the LDM, allowing our method to better resist adversarial guidance.
In the case of DiffAttack (Table \ref{tab:imagenetuaed}), adversarial examples are crafted by inverting latents from the ImageNet validation set, making the resulting attacks more similar to perturbation-based adversarial methods. Despite this difference, our defense continues to achieve state-of-the-art performance. The combination of adversarial anti-aliasing and adversarial super-resolution proves highly effective in mitigating adversarial perturbations, thereby delivering superior robust accuracy without sacrificing clean performance.
Overall, these results highlight that our proposed defense pipeline not only addresses the limitations of existing diffusion-based purification methods but also establishes strong robustness against the emerging class of unrestricted adversarial attacks.

\subsection{Time efficiency}

\begin{table}[t]

\vspace{-2mm}
\caption{\textbf{The average time required to purify a single AutoAttack ($\ell_\infty = 8/255$) adversarial example on the CIFAR-10 dataset.} }
\label{tab:time1}
 \centering
{
\resizebox{0.99\columnwidth}{!}{ \begin{tabular}{l c c c}
        \toprule
         Method & Defend Method & Time Cost(s) & Robust Acc(\%) \\
        \midrule
        Nie \textit{et al.}~\cite{nie2022DiffPure} & Diffusion & 0.6 & 71.03  \\
        Wang \textit{et al.}~\cite{wang2022guided} & Guided Diffusion & 2.2 & 71.18 \\
        Lee \textit{et al.}~\cite{lee2023robust} & Diffusion & 1.8 & 70.47 \\
        Song \textit{et al.}~\cite{song2024mimicdiffusion} & Guided Diffusion & 2.6 &  75.45 \\\midrule
        Ours & Adversarial Anti-Aliasing  & 1e$^{-5}$ &  67.85 \\
         +    & Adversarial Super-Resolution & \textbf{0.07} &  \textbf{82.02} \\
        \bottomrule
    \end{tabular}}}
    
\end{table}

\begin{table}[t]

\vspace{-2mm}
\caption{\textbf{The average time required to purify a single PGD ($\ell_\infty=4/255$) adversarial example on the ImageNet dataset.}}
\label{tab:time2}
 \centering
{
\resizebox{0.99\columnwidth}{!}{\begin{tabular}{l c c c}
        \toprule
         Method & Defend Method & Time Cost(s) & Robust Acc(\%) \\
        \midrule
        Nie \textit{et al.}~\cite{nie2022DiffPure} & Diffusion & 13.3 & 42.88  \\
        Wang \textit{et al.}~\cite{wang2022guided} & Guided Diffusion & 62.8 & 68.78 \\
        Lee \textit{et al.}~\cite{lee2023robust} & Diffusion & 32.4 & 46.31 \\
        Song \textit{et al.}~\cite{song2024mimicdiffusion} & Guided Diffusion & 146.1 &  62.16 \\\midrule
        Ours & Adversarial Anti-Aliasing  & 3e$^{-3}$ &  57.61 \\
         +    & Adversarial Super-Resolution & \textbf{1.1} &  \textbf{69.75} \\
        \bottomrule
    \end{tabular}}}
    
\vspace{-5mm}
\end{table}

We evaluate the average runtime required to defend against a single adversarial example, as reported in Tables \ref{tab:time1} and \ref{tab:time2}. The results show that our proposed method achieves higher robust accuracy while incurring significantly lower computational costs, primarily because it avoids gradient calculations over the diffusion model. In particular, the adversarial anti-aliasing module can neutralize a large proportion of adversarial examples in under one second, making it highly practical for real-time applications. It is also worth noting that the time efficiency of diffusion-based purification methods is largely determined by the resolution of the input images (i.e., the dataset), which can substantially affect scalability.

As previously discussed, employing the entire reverse process with adversarial example guidance is computationally intensive, while using only a partial reverse process diminishes defense performance. In this paper, we propose a two-fold solution to address this issue. First, we introduce an \textbf{effective preprocessing approach, specifically anti-aliasing, to mitigate the impact of adversarial perturbations}. Previous research has shown that diffusion-based adversarial purification should avoid introducing adversarial perturbations into the diffusion model. Therefore, a more effective strategy is to remove some of these perturbations before feeding adversarial examples into the diffusion models. Unlike previous methods that directly utilize adversarial examples for purification, our approach offers a preliminary filtering step.

Second, we employ \textbf{diffusion-based super-resolution instead of diffusion-based image generation}. It is well-known that the reverse process of diffusion models is time-consuming and the gradient calculation exacerbates this issue. However, we may not require the gradient guidance to maintain the semantic consistency of the label between the purified examples and adversarial examples, given that we already have a reference adversarial example. Since adversarial perturbations are pixel-wise, we opt for a relatively lightweight generation task, namely super-resolution, which also focuses on pixel modification. The diffusion-based super-resolution method used in this paper requires only tens of steps, compared to the hundreds or thousands of diffusion steps needed in previous works. With these two approaches, we significantly enhance the time efficiency of diffusion-based adversarial purification without compromising defense performance.

\begin{figure*}[t]
   \centering
     \includegraphics[width=0.7\linewidth]{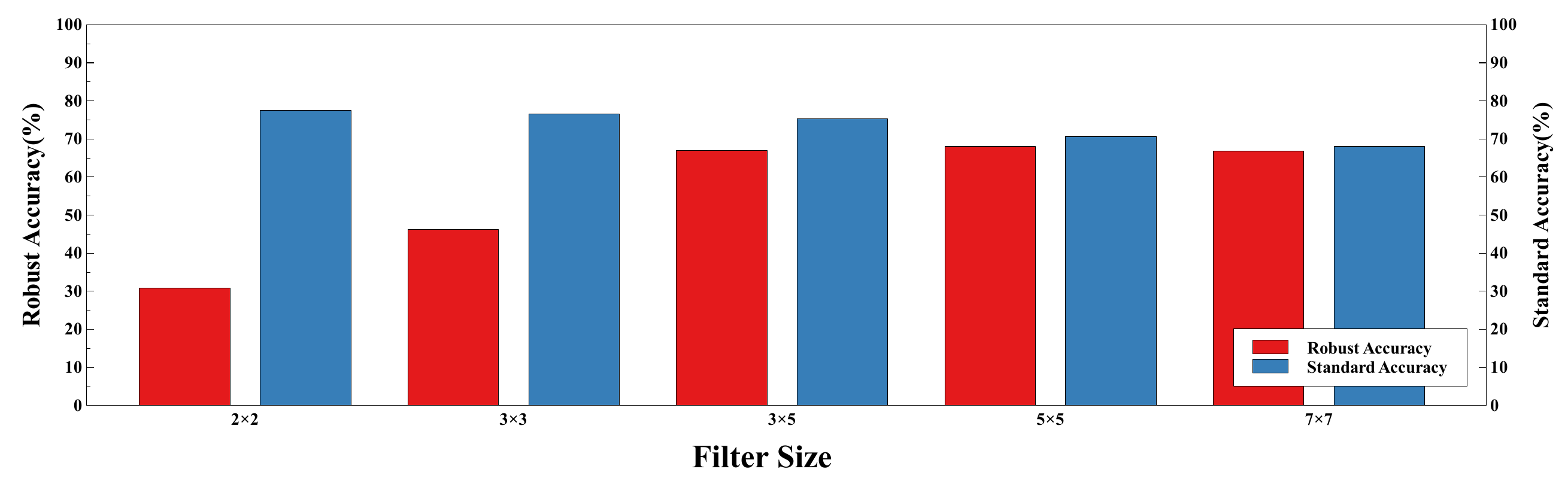}
      \caption{\textbf{The ablation study of filter size.} The weight of the filter at each position is set to 1 except for the center, which we set to 0.}
   \label{fig:ab}
   \end{figure*}

\begin{table*}[t]

\vspace{-2mm}
\caption{\textbf{The defense performance against AutoAttack ($\ell_\infty=8/255$) on the ImageNet dataset.} Time cost on different filters is not evaluated as the computational overhead of adversarial anti-aliasing can be ignored.}

\label{tab:aba}
 \centering
\resizebox{1.5\columnwidth}{!}{  
\begin{tabular}{l c c c c }
        \toprule
         Method & Target Model & Standard Acc(\%) & Robust Acc(\%)  & Time(s)\\
        \midrule
        Ours (Default Settings, 15 steps) & ResNet50 & 75.28 & 67.61  & 1.1\\
        Ours (4 steps) & ResNet50 & 74.81 & 65.30  & \textbf{0.4} \\        
        Ours (50 steps) & ResNet50 & \textbf{76.20} & \textbf{68.13}  & 4.0 \\  \midrule    
        Ours (Default Settings, $3 \times 5$ Mean Filter)& ResNet50 & 75.28 & 67.61  & / \\
        Ours ($5 \times 5$ Mean Filter) & ResNet50& 70.67 & \textbf{68.51}  & / \\
        Ours ($7 \times 7$ Mean Filter)& ResNet50 & 68.31 & 66.81  & / \\    
        Ours ($3 \times 3$ Gaussian Filter) & ResNet50& \textbf{76.21} & 66.89  & / \\
        Ours ($3 \times 5$ Gaussian Filter) & ResNet50& 73.93 & 66.71  & / \\
        Ours ($5 \times 5$ Gaussian Filter) & ResNet50& 71.45 & 66.44  & /

        \\          
        \bottomrule
    \end{tabular}
}
   
\vspace{-2mm}
\end{table*}

\begin{table}[t]
\vspace{-2mm}
\caption{\textbf{The effectiveness of the integration of proposed adversarial purification methods against AutoAttack ($\ell_\infty=8/255$) on the ImageNet dataset.} }
\label{tab:aaa}
 \centering
\resizebox{0.6\columnwidth}{!}{  

   \begin{tabular}{l  c}
        \toprule
         Method & Robust Acc(\%) \\
        \midrule
        Nie \textit{et al.}~\cite{nie2022DiffPure} & 43.89  \\
        + Adversarial AA & 65.28 \\
        + Adversarial AA + SR &  69.44   \\ \midrule
        Song \textit{et al.}~\cite{song2024mimicdiffusion}  &  61.53 \\
        + Adversarial AA & 71.42\\
        + Adversarial AA + SR &   72.18 \\ \midrule
        Adversarial AA   &  55.85 \\
        Adversarial SR  &   41.23 \\
        Adversarial AA+SR  &  \textbf{67.61} \\
        \bottomrule
        
    \end{tabular}
}
   
\vspace{-2mm}
\end{table}

\begin{table}[t]
\vspace{-2mm}
\caption{\textbf{The ablation study on the selection of the super-resolution diffusion model against AutoAttack ($\ell_\infty=8/255$) on the ImageNet dataset.} }
\label{tab:ss}
 \centering
\resizebox{0.8\columnwidth}{!}{  
\begin{tabular}{l c c c}
        \toprule
         Method & Robust Accuracy(\%) & Time(s)\\
        \midrule
        ResShift 256 $\times$ 256 & 67.61 & 1.1 \\
        ResShift 64 $\times$ 64 & 55.98 & 1e$^{-1}$  \\
        LDM-SR 128 $\times$ 128 & 61.35 & 9.0  \\
        SR3  256 $\times$ 256 & 67.24 &  25.1 \\
        \bottomrule
    \end{tabular}
}
   
\vspace{-5mm}
\end{table}

\subsection{Ablation Study}

We perform ablation studies to validate the performance of the proposed methods. We evaluate the defense method against AutoAttack ($\ell_\infty=8/255$) on the ImageNet dataset by default.

\noindent \textbf{Adversarial Anti-Aliasing}. Despite the strong robustness of the proposed adversarial anti-aliasing, the choice of filter configuration plays a critical role in achieving optimal defense performance. Figure \ref{fig:ab} and Table \ref{tab:aba} report results under different filter settings, revealing an inherent trade-off between robust accuracy and standard accuracy. Specifically, robust accuracy tends to stabilize once the filter size exceeds $3\times3$, making it relatively straightforward to identify an appropriate setting with only a few attempts. Moreover, the effectiveness of a chosen filter generalizes across multiple adversarial attack types within the same dataset, as demonstrated in Tables \ref{tab:imagenetaa}, \ref{tab:imagenetpgd}, and \ref{tab:imagenetuae}.
Beyond filter size, the type of filter also substantially influences defense performance. As shown in Table \ref{tab:aba}, while Gaussian filters are commonly used for image degradation in training models such as ResShift, their effectiveness in adversarial purification is limited compared to our proposed mean non-square filter. More critically, Gaussian filters can be directly neutralized by adaptive attacks, further reducing their robustness. In contrast, our mean non-square filter provides a stronger and more resilient defense, highlighting the importance of carefully designing the filtering strategy for adversarial anti-aliasing.

\noindent \textbf{Adversarial Super-Resolution}. The proposed adversarial super-resolution provides a purification mechanism comparable to prior diffusion-based adversarial purification methods, but with the key advantage of avoiding computationally expensive gradient calculations. As shown in Table \ref{tab:aaa}, our method slightly outperforms DiffPure when performing adversarial anti-aliasing to adversarial examples. It is important to note that anti-aliasing is essential for achieving optimal performance, since the current super-resolution process does not explicitly account for adversarial gradients. To address this limitation, we plan to incorporate adversarial training into the super-resolution model in future work.
In addition, we conduct ablation studies with different super-resolution architectures, as reported in Table \ref{tab:ss}. The results highlight the effectiveness of ResShift, further validating its suitability for adversarial purification. Table \ref{tab:aba} shows that increasing the number of diffusion steps improves both standard and robust accuracy. However, this gain comes at the cost of significantly reduced time efficiency due to the longer reverse generation process.

 \noindent \textbf{Adversarial Purification}. We can enhance diffusion-based adversarial purification methods from previous works by replacing the adversarial input with the adversarial examples after the proposed purification. The processed adversarial examples are more benign and closer to the clean images, thereby enabling better purification performance, as shown in Table \ref{tab:aaa}. The performance gain in adversarial purification further validates the necessity of pre-processing adversarial examples before feeding them into the diffusion purification model. Reducing the magnitude of adversarial perturbations enables more effective purification through the denoising-like reverse generation process. However, applying diffusion purification to adversarial examples multiple times does not yield higher robust accuracy, since residual perturbations are beyond the purification threshold.

\noindent \textbf{Deblurring Fine-Tuning}. Our deblurring fine-tuning enhances the performance of the proposed adversarial purification across diverse datasets. While the baseline approach is trained on ImageNet, the fine-tuned method improves both standard and robust accuracy, particularly on CIFAR-10 dataset. This further strengthens the usability and scalability of our defense in practical scenarios across different training dataset.

\section{Limitation}

Despite achieving significantly higher time efficiency and better defense performance than previous diffusion-based adversarial purification methods, our defense approach still has several limitations. One drawback is that the quality of adversarial examples after purification can be relatively low compared to previous purification methods. Therefore, we recommend using the purified images primarily for detecting adversarial inputs. Another limitation is that the robust accuracy against universal adversarial examples (UAEs) is still not on par with perturbation-based adversarial attacks. We aim to address and improve these limitations in future work.

\section{Conclusion}

In this paper, we present an effective and efficient adversarial defense method against both perturbation-based and unrestricted adversarial attacks with theoretical support. The proposed techniques, adversarial anti-aliasing and adversarial super-resolution, effectively eliminate adversarial modifications and recover benign images with minimal computational overhead. The proposed deblurring fine-tuning further enhances the usability and scalability of our purification method across diverse datasets. Comprehensive experiments on the CIFAR-10 and ImageNet datasets validate that our proposed defense outperforms state-of-the-art defense methods. Our work demonstrates that simple adversarial anti-aliasing can achieve moderate model robustness with almost no additional cost. Furthermore, the proposed super-resolution method can perform adversarial purification without requiring the calculation of the diffusion model's gradient. We hope our work will serve as a baseline for the further development of adversarial defenses.

\bibliography{main}
\bibliographystyle{IEEEtran}


\end{document}